\documentclass[letterpaper, 10 pt, conference]{style/ieeeconf}  

\makeatletter
\let\NAT@parse\undefined
\makeatother

\usepackage{dblfloatfix}

\usepackage[numbers,sectionbib,sort&compress]{natbib}
\usepackage{bm}
\usepackage{gensymb}
\usepackage{graphicx}
\usepackage{amsmath}
\usepackage{amssymb}
\usepackage{subcaption}
\usepackage{amsfonts}
\usepackage{siunitx}
\usepackage{makecell}
\usepackage{multirow}
\usepackage{booktabs}
\usepackage{upgreek}
\usepackage[font=small]{caption}
\usepackage[export]{adjustbox}
\usepackage{tikz}
\usepackage{tabularx}
\usepackage{svg}
\usepackage{sidecap} \sidecaptionvpos{figure}{c}
\usepackage{hyperref}
\hypersetup{colorlinks, breaklinks, linkcolor=[rgb]{0.5,0.,0.}, citecolor=[rgb]{0.000,0.427,0.173}, urlcolor=[rgb]{0.031,0.318,0.612}}

\usepackage[nameinlink, capitalize]{cleveref}
\usepackage{color}
\definecolor{CommentPink}{rgb}{1,0.2,0.5}
\definecolor{CommentBlue}{rgb}{0,0,1}
\definecolor{CommentGreen}{rgb}{0,1,0}

\Crefname{section}{Sec.}{Sec.}
\Crefname{equation}{Eq.}{Eq.}
\Crefname{figure}{Fig.}{Fig.}

\DeclareMathOperator*{\argmax}{argmax}

\usepackage[printonlyused,withpage,nolist,nohyperlinks]{acronym}
\begin{acronym}

\acro{2D}{two-dimensional}

\acro{3D}{three-dimensional}

\acro{AHRS}{attitude and heading reference system}

\acro{AUV}{autonomous underwater vehicle}

\acro{CPP}{Chinese Postman Problem}

\acro{DoF}{degree of freedom}
\acrodefplural{DoF}[DoFs]{degrees of freedom}

\acro{DVL}{Doppler velocity log}

\acro{FSM}{finite state machine}

\acro{IMU}{inertial measurement unit}

\acro{LBL}{Long Baseline}

\acro{MCM}{mine countermeasures}

\acro{MDP}{Markov decision process}
\acrodefplural{MDP}[MDPs]{Markov decision processes}

\acro{POMDP}{Partially Observable Markov Decision Process}
\acrodefplural{POMDP}[POMDPs]{Partially Observable Markov Decision Processes}

\acro{PRM}{Probabilistic Roadmap}
\acrodefplural{PRM}[PRM]{Probabilistic Roadmaps}

\acro{ROI}{region of interest}
\acrodefplural{ROI}[ROIs]{regions of interest}

\acro{ROS}{Robot Operating System}

\acro{ROV}{remotely operated vehicle}

\acro{RRT}{Rapidly-exploring Random Tree}
\acrodefplural{RRT}[RRTs]{Rapidly-exploring Random Trees}

\acro{SLAM}{Simultaneous Localisation and Mapping}

\acro{SSE}{sum of squared errors}

\acro{STOMP}{Stochastic Trajectory Optimization for Motion Planning}

\acro{TRN}{Terrain-Relative Navigation}

\acro{UAV}{unmanned aerial vehicle}

\acro{USBL}{Ultra-Short Baseline}

\acro{IPP}{informative path planning}

\acro{FoV}{field of view}
\acrodefplural{FoV}[FoVs]{fields of view}

\acro{CDF}{cumulative distribution function}

\acro{ML}{maximum likelihood}

\acro{RMSE}{Root Mean Squared Error}
\acro{MLL}{Mean Log Loss}

\acro{GP}{Gaussian Process}
\acrodefplural{GP}[GPs]{Gaussian Processes}

\acro{KF}{Kalman Filter}

\acro{IP}{Interior Point}
\acro{BO}{Bayesian Optimization}

\acro{SE}{squared exponential}

\acro{UI}{uncertain input}

\acro{MCL}{Monte Carlo Localisation}
\acro{AMCL}{Adaptive Monte Carlo Localisation}

\acro{SSIM}{Structural Similarity Index}

\acro{MAE}{Mean Absolute Error}
\acro{RMSE}{Root Mean Squared Error}
\acro{AUSE}{Area Under the Sparsification Error curve}

\acro{AL}{active learning}
\acro{DL}{deep learning}
\acro{CNN}{convolutional neural network}

\acro{MC}{Monte-Carlo}

\acro{GSD}{ground sample distance}

\acro{BALD}{Bayesian active learning by disagreement}

\acro{fCNN}{fully convolutional neural network}
\acro{FCN}{fully convolutional neural network}

\acro{CMA-ES}{covariance matrix adaptation evolution strategy}

\acro{FoV}{field of view}

\acro{mIoU}{mean Intersection-over-Union}
\acro{ECE}{expected calibration error}

\acro{GAN}{Generative Adversarial Network}
\acro{POMCP}{Partially Observable Monte-Carlo Planning}
\acro{MCTS}{Monte-Carlo tree search}
\acro{RL}{reinforcement learning}

\end{acronym}

\IEEEoverridecommandlockouts                           

\title{Active Learning of Robot Vision Using Adaptive Path Planning}
\author{\hspace{5mm} Julius R\"{u}ckin \and Federico Magistri \and Cyrill Stachniss \and Marija Popovi\'{c}
\thanks{J.R., F.M., and C.S. are with the Institute of Geodesy and Geoinformation, Cluster of Excellence PhenoRob, University of Bonn. M.P. is with MAVLab, Faculty of Aerospace Engineering, TU Delft. C.S. is also with the University of Oxford, UK, and the Lamarr Institute for Machine Learning and Artificial Intelligence, Germany. This work has been funded by the Deutsche Forschungsgemeinschaft (DFG, German Research Foundation) under Germany's Excellence Strategy, EXC-2070 -- 390732324 (PhenoRob).
Corresponding: \texttt{jrueckin@uni-bonn.de}.}
}

\begin{document}

\maketitle

\begin{abstract}
Robots need robust and flexible vision systems to perceive and reason about their environments beyond geometry. Most of such systems build upon deep learning approaches. As autonomous robots are commonly deployed in initially unknown environments, pre-training on static datasets cannot always capture the variety of domains and limits the robot's vision performance during missions. Recently, self-supervised as well as fully supervised active learning methods emerged to improve robotic vision. These approaches rely on large in-domain pre-training datasets or require substantial human labelling effort. To address these issues, we present a recent adaptive planning framework for efficient training data collection to substantially reduce human labelling requirements in semantic terrain monitoring missions. To this end, we combine high-quality human labels with automatically generated pseudo labels. Experimental results show that the framework reaches segmentation performance close to fully supervised approaches with drastically reduced human labelling effort while outperforming purely self-supervised approaches. We discuss the advantages and limitations of current methods and outline valuable future research avenues towards more robust and flexible robotic vision systems in unknown environments.
\end{abstract}

\section{Introduction} \label{sec:intro}
Perceiving and understanding complex environments is a crucial prerequisite for autonomous systems~\citep{Lenczner2022, Georgakis2021}. In many applications, such as terrain monitoring~\citep{marchant2014sequential, hitz2017adaptive}, search and rescue~\citep{niroui2019deep, baxter2007multi}, and precision agriculture~\citep{popovic2020informative}, autonomous robots need to operate in unknown and unseen environments. This poses a major challenge for classical deep learning-based vision systems, which are trained on static datasets and often do not generalise well to new conditions encountered during real-world deployments.

This work examines the problem of semi-supervised active learning to improve robotic vision within an initially unknown environment while minimising human labelling requirements. We tackle this problem by adaptively re-planning the robot's paths online to collect informative training data to re-train its vision system after a mission. We incorporate two sources of labels for network re-training based on the collected data: (i)~a human annotator and (ii)~automatically generated pseudo labels based on an environment map incrementally built online during a mission.

\begin{figure}[!t]
    \centering
    \includegraphics[width=0.95\columnwidth]{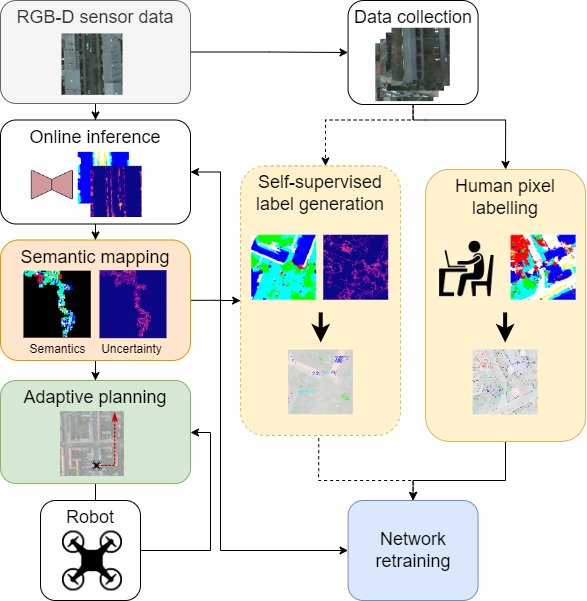}
    \caption{Our semi-supervised active learning approach in an unknown environment. During a mission, a semantic segmentation network predicts pixel-wise semantics and model uncertainties from an RGB-D image. Both are fused into an uncertainty-aware semantic map, which is used by our adaptive planner to guide the robot towards areas of informative training data where model uncertainty is high. After a mission, the collected data is labelled using two sources of labels: (i) human pixel labelling and (ii) self-supervised pseudo label generation from the semantic map.}
    \vspace{-2em}
    \label{F:teaser}
\end{figure}

Active learning is a common approach for reducing human labelling data requirements in computer vision. In the traditional setting, active learning methods select the most informative images from a large, unlabelled dataset~\citep{freund1997selective, gal2017deep, sener2017active, yang2017suggestive}. The selection criterion is commonly derived based on uncertainty, e.g. using Monte-Carlo dropout~\citep{gal2017deep} or ensembles~\citep{beluch2018power}. These approaches are typically not applicable for robot deployments in unknown environments since the collected data is not known in advance. Thus, recent works investigate combining active learning with robotic planning to guide a robot towards parts of the environment with more informative training data for semantic segmentation~\citep{Blum2019, rueckin2022iros, rueckin2023tro}. A drawback of such methods is that the collected images need to be densely labelled, which is still time- and labour-intensive.

Conversely, self-supervised active learning methods automatically generate pseudo labels from maps incrementally built during a mission~\citep{frey2021continual, zurbrugg2022embodied, chaplot2021seal}, without relying on human labelling. However, their applicability to diverse sets of unknown environments is limited since they require large labelled in-domain pre-training datasets to produce high-quality pseudo labels without systematic prediction errors. These pre-training requirements are typically hard to realise in real-world robotic deployment settings, e.g. outdoor and aerial monitoring, where training data is scarce.

Our paper bridges the gap between these two streams of research. We start with a semi-supervised adaptive path planning framework for robotic active learning, introduced in our recent journal publications~\citep{rueckin2024ral, rueckin2023tro}. As illustrated in~\Cref{F:teaser}, the approach combines automatically generating uncertainty-aware self-supervised pseudo labels from a semantic map and selecting informative human-labelled training data. We explore sparse human label selection techniques to further reduce labelling requirements~\citep{shin2021all, xie2022towards}. For adaptive planning, our approach maintains an uncertainty-aware semantic map, enabling us to guide the robot to collect images for labelling from high-uncertainty areas. By combining human and pseudo labels, our goal is to maximise semantic segmentation performance while reducing human labelling effort compared to previous fully supervised works in robotic active learning. Based on our findings, our paper concludes with a new and previously unpublished discussion of limitations and open challenges to drive the research community supporting label-efficient robotic learning paradigms.

\section{Our Approach} \label{S:our_approach}

Considering a robot equipped with an RGB-D sensor, we present an approach for collecting images in an unknown environment to improve semantic segmentation with minimal human labelling effort~\citep{rueckin2023tro, rueckin2024ral} as depicted in \cref{F:teaser}.

\subsection{Probabilistic Semantic Environment Mapping} \label{subsec:mapping}

A crucial requirement for pseudo label generation and adaptive planning is a probabilistic map capturing information about the environment. We use probabilistic multi-layered semantic environment mapping to fuse semantic model predictions. The environment is discretised into two voxel maps $\mathcal{M}_{S}: V \to \{0,1\}^{K \times W \times L \times H}$ and $\mathcal{M}_{U}: V \to [0,1]^{W \times L \times H}$ defined over $W \times L \times H$ spatially independent voxels $V$. The semantic map $\mathcal{M}_{S}$ consists of $K$ layers with one layer per class and is recursively updated using occupancy grid mapping~\citep{Elfes1989}. The model uncertainty map $\mathcal{M}_{U}$ is updated using maximum likelihood estimation. Additionally, we maintain a count map $\mathcal{M}_{T}: V \to \mathbb{N}^{W \times L \times H}$ to track the occurrences in the human-labelled training data utilised in our planning objective. The semantic predictions and model uncertainties change as the semantic segmentation model is re-trained after each robot mission. Thus, we re-compute the semantic and model uncertainty maps after model re-training using previously collected RGB-D images to obtain maximally up-to-date map priors for adaptive planning.

\subsection{Adaptive Informative Path Planning} \label{subsec:planning}

We aim to maximise the performance of a semantic segmentation model with minimal human labelling effort after re-training it on the collected training data. Our map-based global planning methods search for a path $\psi^* = (\mathbf{p}_1,\ldots,\mathbf{p}_N) \in \Psi$ with a variable number $N \in \mathbb{N}$ of robot poses $\mathbf{p}_i \in \mathbb{R}^D$, $i \in \{1,\ldots,N\}$, in the set of potential paths $\Psi$, that maximises an information criterion $I: \Psi \to \mathbb{R}_{\geq 0}$:
\begin{equation} \label{eq:ipp_problem}
    \psi^* = \argmax_{\psi \in \Psi} I(\psi),\, \mathrm{s.t.}\, C(\psi) \leq B\,,
\end{equation}
where $I$ assigns an information value to each possible path $\psi \in \Psi$, $B \geq 0$ is the mission budget, and $C: \Psi \to \mathbb{R}_{\geq 0}$ defines the required budget to execute the path $\psi$.

At each time step $t$, we adaptively re-plan the path $\psi^*_t$ based on the current map states $\mathcal{M}^t_U$ and $\mathcal{M}^t_T$, and execute the next-best pose $\mathbf{p}_{t+1}$ to collect informative training data. The information criterion estimates the effect of a candidate training image recorded at pose $\mathbf{p}$ on a semantic segmentation model's performance. To this end, our information criterion $I$ trades off between model uncertainty and training data diversity. Based on the camera's field of view, we compute a set of voxels $V_{\mathbf{p}}$ visible from pose $\mathbf{p}$ and extract currently mapped model uncertainties $\mathcal{M}^t_U(v)$ and training data occurrences $\mathcal{M}^t_T(v)$ for all voxels $v \in V_{\mathbf{p}}$. Pose~$\mathbf{p}$ contains high information value if model uncertainties $\mathcal{M}^t_U(v)$ are high while training data occurrences $\mathcal{M}^t_T(v)$ are low. To foster exploration, voxels $v$ in unknown space receive a constant exploration bonus $\mathcal{M}^t_U(v) = c_u$, where $c_u > 0$. 

\subsection{Efficient Labelling} \label{subsec:ssl_network_training}

We propose a semi-supervised training strategy for improving the robot's semantic vision. We utilise a semantic segmentation network to predict the pixel-wise probabilistic semantic labels of images. To maximise model performance, we combine human-labelled and automatically pseudo-labelled images during network training.

Combining ideas from \citet{shin2021all} and \citet{xie2022towards}, we propose a new model architecture-agnostic pixel selection procedure for sparse human labels that trades off between label informativeness and diversity. After each mission, we predict each pixel's maximum likelihood semantic label and compute its region impurity score following \citet{xie2022towards}. A pixel's region impurity and, thus, its information value upon re-training is high whenever the number of different classes predicted within its neighbourhood is high, as semantics are usually locally non-cluttered. We select the $\beta\,\%$ pixels with the highest region impurity to ensure an information value lower bound. Then, we sample $\alpha$ pixels uniformly at random from these $\beta\,\%$ pixels to foster training data diversity.

Similarly to self-supervised robotic active learning approaches~\cite{frey2021continual, zurbrugg2022embodied}, we use our incrementally online-built semantic and model uncertainty maps (\Cref{subsec:mapping}) to generate pseudo labels. Given a pose, we render pixel-wise maximum likelihood semantic pseudo labels and model uncertainties from these maps. After each mission, for all images collected in any of the previous missions from respective robot poses, we (re-)render pseudo labels and model uncertainties based on the most recent map beliefs. In contrast to previous works~\citep{frey2021continual, zurbrugg2022embodied}, we only use a sparse set of $\alpha$ pseudo-labelled pixels per image as we experimentally found that sparse pseudo labels balance the human and self-supervision best. Building upon \citet{shin2021all}, for each image, we select the $\beta \%$ pixels with the lowest map-based model uncertainties to ensure a lower bound on the pseudo label quality. Then, we sample $\alpha$ pixels uniformly at random from these $\beta\,\%$ pixels to foster diversity of the sparse pseudo labels.

\section{Experimental Results} \label{sec:exp_eval}

\begin{figure}[!t]
    \centering
    \includegraphics[width=\columnwidth]{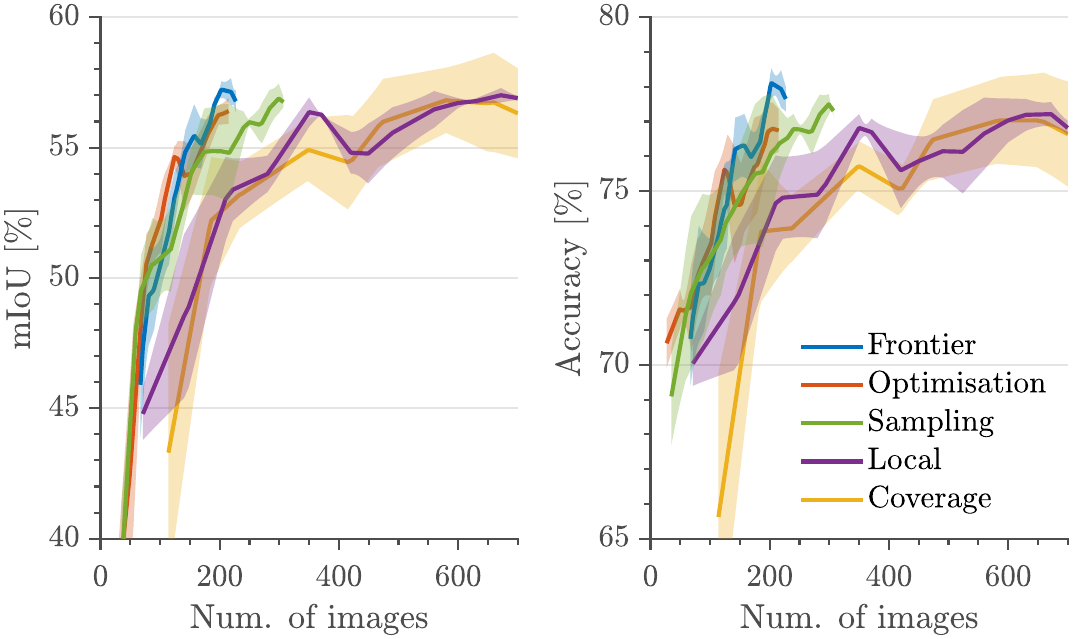}
    \caption{Our global map-based adaptive planners (blue, orange, green) compared to state-of-the-art local planning (purple) and classical non-adaptive coverage paths (yellow). Our map-based planners require substantially fewer pixel-wise human-labelled images to reach the same performance as coverage and local planning.}
    \label{F:planners}
\end{figure}

We evaluate our framework on the real-world ISPRS Potsdam orthomosaic dataset~\cite{Potsdam2018} and simulate $10$ UAV missions from $30$\,m altitude with a mission budget of $1800$\,s. The UAV uses a downwards-facing RGB-D camera with a footprint of $400$\,px$\times 400$\,px. In this work, we consider four adaptive planners~\citep{rueckin2023tro} to optimise our planning objective proposed in \cref{subsec:planning} and a standard coverage pattern:

\textbf{Local} is an image-based planner locally following the direction of the highest training data information in the image recorded at the current UAV position. This planner resembles the state-of-the-art method by \citet{Blum2019}; 

\textbf{Frontier} is a global map-based geometric planner guiding the UAV towards frontiers of explored and unexplored terrain with the highest training data information; 

\textbf{Optimisation} selects a path over a fixed horizon of multiple time steps to optimise the path's overall training data information following the work by \citet{popovic2020informative};

\textbf{Sampling} utilises \ac{MCTS}~\cite{browne2012survey} to find the next position that maximises the future training data information in a sampling-based fashion.

We use Bayesian ERFNet~\citep{rueckin2022iros} pre-trained on the Cityscapes dataset~\citep{cordts2016cityscapes}. Re-training after each mission starts from this checkpoint and stops after convergence on the validation set. We use a one-cycle learning rate, a batch size of $8$, and weight decay $\lambda = (1 - p) / 2N$, where $p = 0.5$ is the dropout probability and $N$ is the number of images~\cite{gal2017deep}.

In~\Cref{F:planners}, we evaluate the performance of the adaptive planners against a traditional pre-planned coverage-based training data collection. We report the \ac{mIoU} and accuracy over the number of pixel-wise human-labelled training images averaged over three different UAV starting locations. Higher semantic segmentation performance, thanks to newly added images, indicates better active learning and, thus, planning performance. The local planner, on average, does not perform better than the coverage baseline. All our adaptive map-based planners, on average, reach higher active learning performance than the coverage baseline (yellow) and local planner (purple) with substantially fewer human-labelled images. Specifically, the frontier planner (blue) requires approx. $200$ images to reach the performances of the coverage planner on approx. $600$ images. These results verify that our global map-based adaptive planners outperform classical pre-planned data collection campaigns for active learning in semantic terrain mapping missions. Further, they show that our map-based adaptive planners reach higher active learning performance than previous state-of-the-art local planning~\citep{Blum2019}.

\begin{figure}[!t]
    \centering
    \includegraphics[width=\columnwidth]{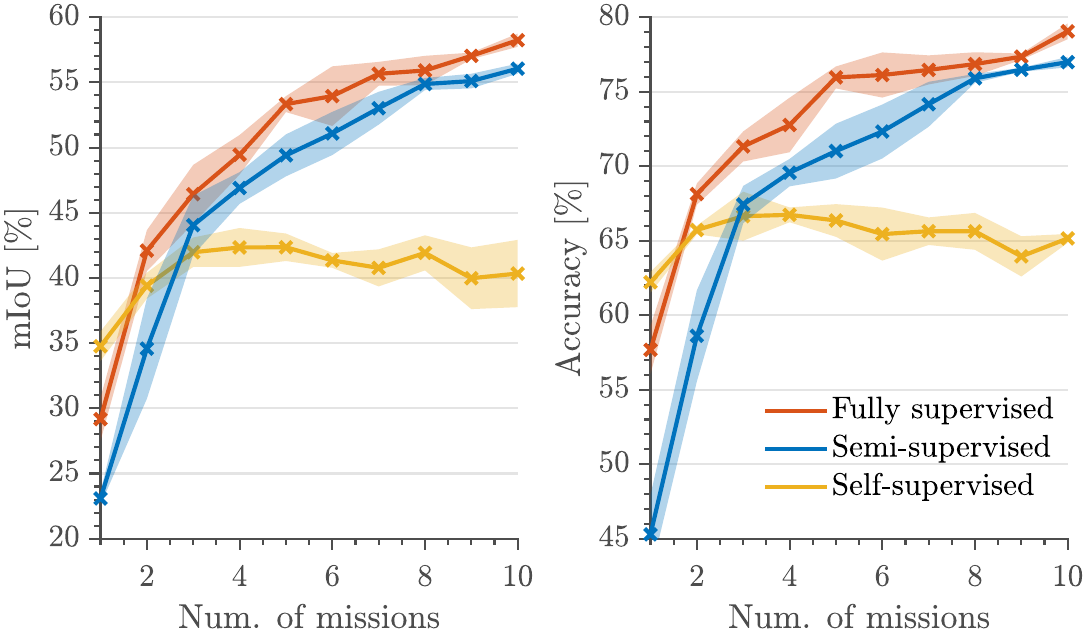}
    \caption{Our semi-supervised adaptive frontier planning compared to fully and self-supervised adaptive frontier planning. Our semi-supervised approach almost reaches the fully supervised performance while clearly outperforming the self-supervised approach.}
    \label{F:results}
\end{figure}

In~\Cref{F:results}, we select the map-based adaptive frontier planner to evaluate the effect of our proposed semi-supervised training data labelling strategy. We report the \ac{mIoU} and accuracy after each mission's network re-training averaged over three different runs to account for the inherent randomness in the sparse pixel selection procedure. We compare our semi-supervised labelling strategy (blue) to (i) fully supervised pixel-wise human labels (orange), and (ii) to a purely self-supervised labelling strategy (yellow) pre-trained on a small set of human-labelled Potsdam ISPRS images and rendering pixel-wise pseudo labels based on the current semantic map belief. Our semi-supervised strategy performs almost on par with the fully supervised human labelling while requiring approx. only $0.5\%$ of the human-labelled pixels. Interestingly, the self-supervised approach fails to improve model performance after four re-deployments. This indicates that efficiently selecting human-labelled pixels is a key ingredient of our framework to circumvent reinforced self-supervision errors in semantic terrain mapping missions.

\section{Discussion \& Future Directions} \label{sec:discussion}

Next, we discuss the open challenges in adaptive robotic planning for active learning of robust vision and suggest future research directions to address them.

\subsection{Faster to Answer Human Labelling Queries} \label{subsec:discussion_fast_labelling}

Although self-supervised methods do not require human annotations to improve vision performance, these approaches rely on large human-labelled pre-training datasets containing data similar to deployment. Thus, self-supervised methods are often upper-bounded in performance by the pre-training and domain shift during deployment. In contrast, fully supervised methods induce substantial human labelling costs requiring pixel-wise annotations~\citep{Blum2019, rueckin2022iros, rueckin2023tro}. Our results show that, for active learning in semantic terrain mapping, combining sparse human labels and self-supervision enables reducing the number of human-labelled pixels to approx. $0.5\%$ of fully supervised methods while maintaining performance~\citep{rueckin2024ral}. Although recent studies suggest that sparse pixel selection reduces annotation time~\citep{shin2021all, benenson2022colouring}, human labelling query costs are, from our perspective, still too high to be easily and repeatedly answered by an operator. One idea could be to explore uncertainty-guided one-click annotations~\citep{xu2016deep, majumder2020multi}. Another promising path could be to leverage foundation models, such as SAM~\citep{Kirillov2023sam}, and prompt them in a targeted, potentially uncertainty-aware fashion. 

\subsection{Novel Embodied Self-supervised Learning Methods}

Human-guided methods still suffer from costly human annotations~\citep{Blum2019, rueckin2023tro, rueckin2024ral}. High-quality self-supervised labels are required to keep the human labelling effort low and reach maximal prediction performance. To this end, self-supervised methods create pseudo labels from an online-built semantic map~\citep{chaplot2021seal, frey2021continual, zurbrugg2022embodied, rueckin2024ral}. These methods render pseudo labels from voxel-based maps at viewpoints encountered during deployment. However, voxel-based maps cannot render image-label pairs from novel viewpoints. Semantic neural rendering approaches recently enhanced self-supervised pseudo labels, rendering high-quality image-label pairs from novel viewpoints, outperforming voxel map-based pseudo label generation~\citep{liu2023unsupervised}. Combining neural rending methods with adaptive planning could improve current systems without additional human labels. Further, robotic active learning methods leverage the robot's embodiment in the environment using adaptive planning to enhance spatial consistency of pseudo labels~\citep{zurbrugg2022embodied, chaplot2021seal}. Most methods use generated map-based pseudo labels directly with standard loss functions during network training~\citep{frey2021continual, zurbrugg2022embodied}, but they do not leverage advanced self-supervised methods, such as contrastive learning. \citet{chen2024embodied} show that additionally enforcing spatial consistency during network training using contrastive learning techniques improves object-goal navigation. These advanced self-supervised techniques could also improve active learning for robotic vision systems. 

\subsection{Improved Uncertainty Quantification} \label{subsec:discussion_uncertainty_quantification}

As discussed by \citet{chaplot2021seal}, overconfidently wrong predictions reinforce prediction errors after re-training on these predictions in a self-supervised fashion. Even human-guided methods require well-calibrated uncertainty estimation to create informative human labelling queries that maximise performance while minimising labelling effort~\citep{rueckin2023tro}. Thus, better-calibrated model uncertainty estimation techniques are required as current techniques tend to produce overconfident predictions~\citep{postels2021practicality, gal2016dropout, beluch2018power}. Further, current methods ignore various sources of uncertainty. All methods use some measure of model uncertainty or confidence~\citep{zurbrugg2022embodied, rueckin2023tro, rueckin2024ral, chaplot2021seal} to collect potentially informative new training data. Future research could integrate and disentangle other sources of uncertainty, such as data uncertainty~\citep{kendall2017uncertainties} induced by environmental factors or noisy sensors. This information could be used to avoid requesting human labels for inputs with high data uncertainty that contribute little to the model improvements~\citep{kendall2017uncertainties} or to adaptively plan novel viewpoints that might reduce these uncertainties~\citep{morilla2024perceptual}.

\subsection{Towards Continual Active Learning} \label{subsec:discussion_continual_learning}


Another key challenge for efficient learning of robotic vision systems is the robot's ability to continually learn about new unseen environments while transferring the knowledge gained during previous deployments~\citep{lesort2020continual} without suffering from catastrophic forgetting~\citep{mccloskey1989catastrophic}. This problem of continual learning is largely ignored in robotic active learning methods. To the best of our knowledge, \citet{frey2021continual} proposed the only method for continual active learning using experience replay~\citep{rolnick2019experience}. However, they do not leverage adaptive planning for training data collection. Further, although conceptually simple and effective against catastrophic forgetting, experience replay is storage- and compute-inefficient as its complexity scales linearly with the number of deployments. Combining adaptive replanning with continual learning over sequential deployments in various environments could lead to more robust vision systems and a more targeted continuous collection of informative training data while leveraging already gained previous knowledge.

\subsection{Improved Model Re-training Efficiency} \label{subsec:discussion_efficient_retraining}


Similarly to continual active learning, current methods for active learning within a single environment require iterative network re-training to adapt training data collection based on previously collected data. Although most methods use lightweight networks for improved training and inference speed~\citep{zurbrugg2022embodied, rueckin2023tro, rueckin2024ral}, iterative re-training is prohibitively expensive in applications that require fast online adaption of vision or re-deployment cycles. One way to improve the network re-training efficiency could be to leverage vision foundation models~\citep{Kirillov2023sam} as pre-trained feature extractors combined with small, trainable adapter networks. This could mitigate the costly re-training of larger networks while allowing the robot vision to profit from few-shot generalisation.

\section{Conclusion} \label{S:conclusions}

We presented our adaptive planning approach for semi-supervised active learning of robotic vision in unknown environments~\citep{rueckin2024ral}. 
Our experimental results show that our semi-supervised approach outperforms traditional pre-planned data collection campaigns and purely self-supervised robotic active learning approaches in semantic terrain monitoring missions. Further, our approach requires only approx. $0.5 \%$ of the human-labelled pixels of fully supervised robotic active learning methods~\citep{rueckin2023tro} while maintaining semantic segmentation performance. We conclude with a discussion of open challenges and identify future directions to advance state-of-the-art robotic active learning methods.

\bibliographystyle{IEEEtranN}
\footnotesize
\bibliography{2024-iros-bob-workshop-rueckin}

\end{document}